\documentclass[cameraready]{Interspeech}

\title{InfoShield: Privacy-Preserving Speech Representations for Mental Health Screening via Information-Theoretic Optimization}

\author[affiliation={1},correspondingauthor]{Xueyang}{Wu}
\author[affiliation={2}]{Siyuan}{Liu}
\author[affiliation={3}]{Kezhuo}{Yang}
\author[affiliation={1}]{Guang}{Ling}

\address{
    $^1$ Shenzhen NeurStar Inc., China \\
    $^2$ University of York, United Kingdom \\
    $^3$ Shanghai Jiao Tong University, China
}

\email{wuxueyang@neurstar.ai, sl1547@york.ac.uk, ykz@sjtu.edu.cn, lingguang@neurstar.ai}


\def\R{{\mathbb R}}

\usepackage{tikz}
\usetikzlibrary{positioning, arrows.meta, shapes.geometric, calc}

\begin{document}

\maketitle

\begin{abstract}
Speech-based mental health screening offers scalable depression detection, yet clinical deployment faces a significant barrier: users' privacy concerns about demographic information exposure. Current techniques struggle to resolve this conflict. Adversarial training often fails against unseen threats, whereas Differential Privacy tends to compromise diagnostic performance by injecting noise across all features. This paper presents InfoShield, which minimizes mutual information between speech representations and sensitive attributes while preserving depression classification accuracy. We identify that standard MINE estimators struggle with sequential speech due to temporal-static misalignment, and introduce TimeAwareMINE with cross-modal attention to align acoustic frames with attribute embeddings. Experiments on the Androids Corpus show InfoShield reduces gender inference from 92.6\% to 55.5\% and age inference from 55.7\% to 30.3\% with limited utility loss (6\% F1 reduction), achieving F1=0.784 compared to prior SOTA's 0.723~\cite{alsarrani2022thin}.
\end{abstract}

\section{Introduction}
\label{sec:intro}

Depression affects approximately 4.4\% of the global population~\cite{who2022mental}. Speech-based screening offers scalable, non-invasive depression detection as acoustic features encode clinically relevant biomarkers~\cite{cummins2018speech,low2020voice}. However, clinical deployment faces a significant barrier: users' privacy concerns about demographic information exposure. Studies show privacy worries deter adoption, as employers, insurers, or governments might infer protected attributes like gender, age, or socioeconomic status from voice recordings~\cite{tao2023androids}. The patients who could benefit most from screening may be least willing to share speech data.

\textbf{Problem Statement.} Speech signals inherently encode both diagnostic biomarkers and sensitive demographic traits. This implies that standard acoustic representations inevitably leak private information alongside depression cues (see Section~\ref{sec:experiments}). Balancing diagnostic precision with privacy is therefore challenging. Current solutions are ill-suited for this task: adversarial training offers protection only against anticipated attacks~\cite{fan2019adv}, whereas Differential Privacy (DP)~\cite{dwork2006differential} applies noise uniformly, often degrading the fine-grained spectral patterns required for screening. Crucially, neither method explicitly measures or targets the removal of sensitive attributes from the latent space.

\textbf{Our Approach.} We propose InfoShield to secure speech representations in mental health tools. The framework integrates Variational Information Bottleneck (VIB) compression for better generalization with targeted mutual information (MI) minimization for privacy. By optimizing these objectives simultaneously—balancing $\text{KL}[q_\phi(Z|X) \Vert p(Z)]$ against $\hat{I}(Z;s)$—the model learns to retain diagnostic markers while suppressing sensitive attributes.

\textbf{Key Limitation Identified.} Standard MINE estimators struggle with sequential speech due to temporal-static misalignment between time-varying acoustics and static attribute labels, leading to unreliable MI estimates. We introduce TimeAwareMINE with cross-modal attention to align acoustic frames with attribute representations.

This paper makes three contributions:
\begin{itemize}
    \item \textbf{TimeAwareMINE}: Standard MINE pools variable-length acoustic sequences but receives a single static attribute label, causing temporal-static misalignment. Our cross-modal attention solution achieves better utility (F1: 0.782 vs. 0.714) and age privacy (39.7\% vs. 43.5\%) compared to StandardMINE.
    \item \textbf{Unified Framework}: Integrating VIB compression with TimeAwareMINE achieves better privacy-utility balance than individual components (Gender: 55.5\%, Age: 30.3\%, F1: 0.784).
    \item \textbf{Evaluation on Androids Corpus}: Ablation studies show InfoShield reduces gender inference from 92.6\% to 55.5\% and age inference from 55.7\% to 30.3\% with 6\% utility loss, outperforming DP baselines and prior SOTA~\cite{alsarrani2022thin} (F1: 0.784 vs. 0.723).
\end{itemize}

\section{Related Work}
\label{sec:related}

\textbf{Speech-Based Mental Health Screening.} Speech analysis demonstrates significant potential for non-invasive depression detection~\cite{cummins2018speech,low2020voice}. Recent deep learning approaches achieve competitive performance~\cite{alsarrani2022thin,tao2023androids}, yet clinical deployment remains limited due to privacy concerns about demographic leakage~\cite{de2024probing}. Our work addresses this gap with privacy-preserving representations specifically designed for mental health applications.

Existing speech privacy approaches fall into three categories, each with limitations for demographic attribute protection:

\textbf{Traditional Privacy Methods} focus on speaker anonymization~\cite{snyder2017deep,fang2019speaker} or adversarial training against specific attackers. These methods lack generalizability---they protect against known adversaries but fail against novel attack architectures. In contrast, our information-theoretic approach provides universal protection without requiring knowledge of specific attack models.

\textbf{Differential Privacy} provides formal guarantees through global noise injection~\cite{abadi2016deep,pelikan2023federated}, but this affects all learned features indiscriminately, including diagnostically relevant acoustic patterns. Our targeted MI minimization selectively removes sensitive information while preserving task-relevant signals through principled VIB compression.

\textbf{Information-Theoretic Approaches} offer principled representation learning through the Information Bottleneck~\cite{tishby2000information} and MINE~\cite{belghazi2018mine}. Recent advances include CLUB~\cite{cheng2020club} for tighter MI bounds and MI-based speaker learning~\cite{ravanelli2019learning}. However, standard MINE suffers from high variance on sequential data~\cite{poole2019variational}. Recent work targets different threats: USC~\cite{vecino2025universal} protects speaker identity, SafeEar~\cite{safeear2024} prevents deepfake detection---none address comprehensive privacy protection in clinical speech analysis. Our framework integrates VIB compression with TimeAwareMINE for joint optimization of utility, privacy, and generalization.

\section{Methodology}
\label{sec:method}

\subsection{InfoShield Architecture}

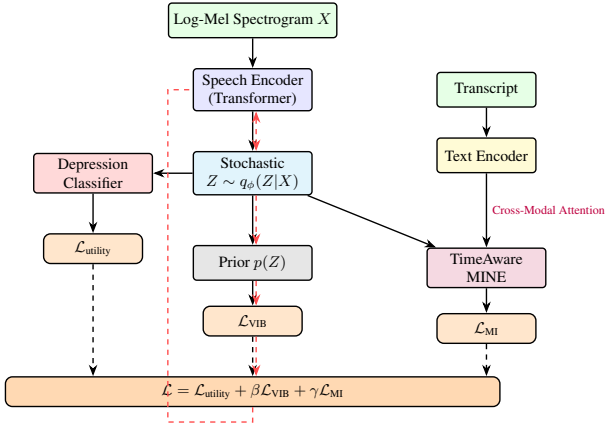
\begin{figure}[t]
\centering
\resizebox{\columnwidth}{!}{%
\begin{tikzpicture}[
    node distance=0.8cm and 1.2cm, 
    box/.style={rectangle, draw=black, thick, rounded corners=2pt, minimum height=0.7cm, minimum width=2.4cm, align=center, font=\small},
    loss/.style={rectangle, draw=black, thick, rounded corners=4pt, fill=orange!20, minimum height=0.6cm, minimum width=2cm, align=center, font=\small},
    input/.style={box, fill=green!10},
    encoder/.style={box, fill=blue!10},
    z_node/.style={box, fill=cyan!10},
    classifier/.style={box, fill=red!15},
    prior/.style={box, fill=gray!20},
    mine/.style={box, fill=purple!15},
    text_input/.style={box, fill=green!10, minimum width=2cm},
    text_enc/.style={box, fill=yellow!20, minimum width=2cm},
    arrow/.style={->, >=Stealth, thick, color=black},
    red_arrow/.style={->, >=Stealth, thick, dashed, color=red!70},
    red_line/.style={-, dashed, thick, color=red!70}
]
%
%
\node[input] (input) {Log-Mel Spectrogram $X$};
%
\node[encoder, below=0.6cm of input] (speech_enc) {Speech Encoder\\(Transformer)};
%
\node[z_node, below=0.8cm of speech_enc] (z) {Stochastic\\$Z \sim q_\phi(Z|X)$};
%
\node[prior, below=1.0cm of z] (prior) {Prior $p(Z)$};
%
\node[loss, below=0.5cm of prior] (lvib) {$\mathcal{L}_{\text{VIB}}$};
%
%
\node[classifier, left=0.8cm of z] (classifier) {Depression\\Classifier};
%
\node[loss, below=0.8cm of classifier] (lutility) {$\mathcal{L}_{\text{utility}}$};
%
%
\node[text_input, right=2.5cm of speech_enc] (transcript) {Transcript};
%
\node[text_enc, below=0.6cm of transcript] (text_enc) {Text Encoder};
%
\node[mine, below=1.5cm of text_enc] (tamine) {TimeAware\\MINE};
%
\node[loss, below=0.5cm of tamine] (lmi) {$\mathcal{L}_{\text{MI}}$};
%
%
\node[loss, below=0.8cm of lvib, minimum width=10cm, fill=orange!30] (total) {$\mathcal{L} = \mathcal{L}_{\text{utility}} + \beta\mathcal{L}_{\text{VIB}} + \gamma\mathcal{L}_{\text{MI}}$};
%
%
\draw[arrow] (input) -- (speech_enc);
\draw[arrow] (speech_enc) -- (z);
\draw[arrow] (z) -- (prior);
\draw[arrow] (prior) -- (lvib);
%
\draw[arrow] (z) -- (classifier);
\draw[arrow] (classifier) -- (lutility);
%
\draw[arrow] (transcript) -- (text_enc);
\draw[arrow] (text_enc) -- node[midway, right, font=\scriptsize, text=purple] {Cross-Modal Attention} (tamine);
%
\draw[arrow] (z) -- (tamine);
%
\draw[arrow] (tamine) -- (lmi);
%
\draw[arrow, dashed] (lutility) -- (total.north -| lutility);
\draw[arrow, dashed] (lvib) -- (total);
\draw[arrow, dashed] (lmi) -- (total.north -| lmi);
%
%
\draw[red_arrow, <->, transform canvas={xshift=2pt}] (speech_enc.south) -- (z.north);
\draw[red_arrow, transform canvas={xshift=2pt}] (z.south) -- (prior.north);
\draw[red_arrow, transform canvas={xshift=2pt}] (prior.south) -- (lvib.north);
\draw[red_arrow, transform canvas={xshift=2pt}] (lvib.south) -- (total.north);

\draw[red_line] (total.south) -- ++(0,-0.3) -| (input.west |- speech_enc.west) -- (speech_enc.west);
\end{tikzpicture}%
}
\caption{InfoShield architecture. Left: speech encoder produces latent representations $Z$ for depression classification. Right: TimeAwareMINE quantifies privacy leakage via cross-modal attention with transcript embeddings. Three loss terms jointly optimize the framework. Red dashed arrows indicate gradient backpropagation.}
\label{fig:architecture}
\end{figure}

Figure~\ref{fig:architecture} illustrates the overall InfoShield framework. The input log-mel spectrogram $X$ is processed by a Transformer encoder to produce stochastic latent representations $Z$. Three loss terms jointly optimize the network: (1) utility loss $\mathcal{L}_{\text{utility}}$ for depression prediction, (2) VIB compression loss $\mathcal{L}_{\text{VIB}}$ for regularization, and (3) privacy loss $\mathcal{L}_{\text{MI}}$ via TimeAwareMINE for minimizing information leakage about sensitive attributes.

\subsection{Notation and Problem Formulation}

\textbf{Notation.} We define the following notation used throughout this paper:
\begin{itemize}
    \item $X \in \R^{T \times D}$: input log-mel spectrogram sequence with $T$ frames
    \item $Y \in \{0, 1\}$: binary depression label (target task)
    \item $Z \in \R^{T \times d}$: learned latent representation
    \item $s$: transcript sentence (text input), encoded by BERT for cross-modal attention
    \item $\mathbf{e}_s \in \R^{d'}$: BERT embedding of transcript sentence $s$
    \item $s'$: negative transcript samples drawn from marginal distribution for contrastive MI estimation
\end{itemize}


\subsection{Information-Theoretic Privacy Framework}

\textbf{Problem Formulation.} We formulate privacy-preserving speech representation learning as a constrained optimization problem that directly minimizes mutual information between learned representations and sensitive attributes while preserving diagnostic utility.

\textbf{Objective Function.} Our encoder $q_\phi(Z|X)$ learns stochastic representations $Z$ by optimizing:
\begin{align}
\mathcal{L} = -\mathbb{E}[\log p_\theta(Y|Z)] + \beta \text{KL}[q_\phi(Z|X) \Vert p(Z)] + \gamma \hat{I}(Z;s)
\label{eq:objective}
\end{align}
where the VIB compression term $\text{KL}[q_\phi(Z|X) \Vert p(Z)]$ provides generalization regularization, and $\hat{I}(Z;s)$ explicitly minimizes mutual information between speech representations $Z$ and transcript sentences $s$ containing demographic cues, thereby protecting sensitive attributes from inference attacks.

\textbf{Theoretical Foundation.} The privacy-utility relationship follows from data processing inequalities~\cite{cover1999elements}. For any representation $Z$ derived from input $X$ where $(X,Y,s)$ forms a joint distribution:
\begin{align}
I(Z;Y) \leq I(X;Y) + I(X;s|Y) - I(Z;s)
\label{eq:tradeoff}
\end{align}
This bound reveals the fundamental trade-off: reducing sensitive information $I(Z;s)$ necessarily constrains achievable utility $I(Z;Y)$, assuming the Markov condition $Z \leftarrow X \rightarrow (Y,s)$ holds for our encoder.

\subsection{TimeAwareMINE for Sequential Speech}

\textbf{Mechanism of Protection.} Speech inherently links acoustic features (e.g., pitch) with linguistic patterns. By minimizing mutual information with the transcript $I(Z; s)$, InfoShield forces the model to discard not just linguistic content, but also the statistically correlated acoustic cues. This effectively strips away gender and age information implicit in the signal, blocking attribute inference.

\textbf{Problem Identification.} Standard MINE estimators suffer from estimation challenges on sequential speech: temporal pooling destroys critical time-dependency information, while random temporal-static pairing introduces noise that leads to loose, unreliable MI bounds. This estimation inaccuracy directly undermines privacy optimization effectiveness.

\textbf{Cross-Modal Temporal Alignment.} Given acoustic sequence $Z = (z_1, \ldots, z_T)$ and transcript embedding $\mathbf{e}_s$ from BERT encoding, we compute frame-wise alignment through cross-modal attention:
\begin{align}
\alpha_t &= \text{softmax}(z_t^\top \mathbf{e}_s / \sqrt{d}) \\
c_t &= \sum_{j=1}^T \alpha_{tj} z_j
\label{eq:alignment}
\end{align}
This cross-modal attention mechanism dynamically aligns acoustic frames with transcript representations, addressing the temporal-static misalignment that standard MINE cannot handle when pairing sequential speech with static attribute labels.

\textbf{MI Estimation.} The statistics network $T_\psi$ processes aligned features to estimate:
\begin{align}
\hat{I}(Z;s) = \frac{1}{T}\sum_{t=1}^T \left[ T_\psi(c_t, \mathbf{e}_s) - \log \mathbb{E}_{s'} e^{T_\psi(c_t, \mathbf{e}_{s'})} \right]
\label{eq:mine}
\end{align}

This temporal-aware design produces significantly tighter MI bounds by ensuring meaningful frame-transcript correspondence, leading to improved privacy protection through more accurate information leakage quantification. The MI minimization objective $\hat{I}(Z;s)$ learns representations that reduce mutual information between speech $Z$ and transcripts $s$ containing demographic cues, thereby protecting sensitive attributes (gender, age) from inference attacks.

\textbf{Connection to Attribute Privacy.} Let $A$ denote the sensitive attribute (gender/age) and $S$ denote the transcript. From the chain rule of mutual information and data processing inequality:
\begin{align}
I(Z; A) \leq I(Z; S) + I(A; S | Z)
\label{eq:privacy_chain}
\end{align}
When we minimize $I(Z; S) \to 0$ and $I(A; S | Z)$ is small (weak conditional dependence), we guarantee $I(Z; A) \to 0$. Our empirical results validate this theoretical mechanism: reducing transcript-speech MI correlates with reduced attribute inference accuracy.

\textbf{Why Cross-Modal Attention Matters.} Standard MINE applies global pooling over the temporal dimension, computing a single MI estimate for the entire utterance. This is problematic for speech because different frames contain varying amounts of privacy-relevant information—vowel-heavy segments encode more gender cues due to pitch, while consonants are less revealing. Global pooling dilutes these frame-wise differences. Our cross-modal attention enables frame-wise privacy quantification where the attention weight $\alpha_t$ indicates how much demographic information each frame contains, allowing targeted privacy removal rather than uniform blurring.

\textbf{Theoretical Guarantees.} Under standard regularity conditions (finite VC-dimension for $T_\psi$, $\beta$-mixing temporal dependence), our estimator converges almost surely: $\lim_{n \to \infty} \hat{I}_{TA,n} = I(Z;s)$. We ensure stable training via spectral normalization on the statistics network and exponential moving average of the MI estimate ($\tau=0.99$).

\section{Experiments}
\label{sec:experiments}

\subsection{Experimental Setup}

\textbf{Clinical Dataset:} The Androids Corpus~\cite{tao2023androids} contains 228 recordings from 118 native Italian speakers (64 clinically diagnosed with depression, 54 healthy controls). Similar to prior work~\cite{alsarrani2022thin}, we use interview speech for ecological validity. For privacy evaluation, we extract gender and age groups\footnote{Age groups: Young $\leq$30, Middle 31--45, Senior $\geq$46.} from clinical metadata, simulating privacy-sensitive healthcare deployment.

\textbf{Baselines and Ablations:} We compare against: (1) \textbf{Normal} (base Transformer without privacy, utility oracle), (2) \textbf{DP ($\varepsilon$=1, 8)} (Differential Privacy), (3) \textbf{VIB-only}, (4) \textbf{StandardMINE} / \textbf{TimeAwareMINE}, and (5) \textbf{InfoShield}. This systematic ablation demonstrates each component's contribution.

\textbf{Evaluation Framework:} Given the modest dataset size (118 speakers), our evaluation focuses on feasibility demonstration. All experiments use 5-fold cross-validation with participant-level splits. We assess: (1) \textit{Privacy-Utility Feasibility}: Can information-theoretic approaches achieve privacy without destroying utility? (2) \textit{Method Comparison}: Targeted MI minimization vs. differential privacy. (3) \textit{Component Analysis}: TimeAwareMINE's temporal adaptations vs. standard MINE.

\textbf{Hyperparameter Configuration:} Preliminary experiments show stable performance with utility:privacy:compression loss ratios of 2:1:1. We set $\beta$=0.001, $\gamma$=0.01.

\textbf{Architecture and Training Details:} Our Transformer encoder processes log-mel features through multi-head self-attention, outputting Gaussian parameters for stochastic representations. TimeAwareMINE uses three fully-connected layers with ReLU activations. We employ curriculum learning with privacy weight increasing linearly from 0.001 to target value over the first 25\% of epochs, with MI estimator updating twice per encoder update.

\subsection{Architecture and Implementation Details}

\textbf{Data Preprocessing.} Raw audio was resampled to 16kHz. Log-mel spectrograms were computed with 80 mel bins, 25ms window, and 10ms stride. No data augmentation was applied due to the small dataset size. Transcripts were encoded using sentence-BERT~\cite{sentencebert}.

\textbf{Encoder.} 4-layer Transformer (8 heads, $d_{\text{model}}$=256, dropout=0.3) outputs Gaussian parameters $\{\mu, \sigma\}$ for a 64-dimensional latent $Z$. Training uses 30 MC samples; inference uses mean $\mu$. The 4-layer design balances capacity with overfitting risk on 118 speakers.

\textbf{TimeAwareMINE.} 3-layer FC network (256$\to$128$\to$64) with spectral normalization. Cross-modal attention aligns acoustic frames with BERT attribute embeddings (768-dim) to address temporal-static misalignment.

\textbf{Optimization.} AdamW (lr=$10^{-4}$, weight decay=$10^{-5}$), batch size 32, 5 epochs. We set $\beta=10^{-3}$ (VIB) and $\gamma=0.01$ (privacy) based on grid search---lower values insufficient for privacy, higher values cause utility collapse.

\textbf{Differential Privacy.} Opacus DP-SGD with gradient clipping 1.2, $\delta=10^{-5}$. We report $\varepsilon\in\{1.0, 8.0\}$ showing the privacy-utility spectrum.

\textbf{Attack Model.} 3-layer Transformer attacker (8 heads, $d$=80) trained on frozen representations with class-weighted cross-entropy for age groups: Young ($\leq$30), Middle (31--45), Senior ($\geq$46).

\subsection{Feasibility Validation and Ablation Study}

Tables~\ref{tab:utility} and~\ref{tab:privacy} present comprehensive 5-fold cross-validation results demonstrating our unified framework's effectiveness. The Normal baseline (F1=0.834) establishes the utility upper bound but suffers severe privacy vulnerability: 92.6\% gender and 55.7\% age inference accuracy. Our InfoShield framework achieves optimal privacy-utility balance: gender inference drops to 55.5\% and age inference to 30.3\% (below 33.3\% random chance for 3-class classification), while maintaining competitive depression classification (F1=0.784) with only 6\% utility cost compared to the oracle. InfoShield outperforms SOTA approaches~\cite{alsarrani2022thin} (F1: 0.784 vs. 0.723) while providing strong privacy protection, with TimeAwareMINE showing substantial improvements over StandardMINE for both utility and privacy.

\begin{table}[htb]
\centering
\caption{Diagnostic Performance for Depression Classification on Androids Corpus (5-fold CV)}
\label{tab:utility}
{\footnotesize
\begin{tabular}{lcccc}
\hline
Method & F1 & Accuracy & Precision & Recall \\
\hline
Previous SOTA~\cite{alsarrani2022thin} & 0.723 & 0.676 & 0.717 & 0.722 \\
\hline
Normal (Oracle) & 0.834 & 0.818 & 0.795 & 0.883 \\
 & $\pm$0.105 & $\pm$0.096 & $\pm$0.135 & $\pm$0.107 \\
DP ($\varepsilon$=1) & 0.568 & 0.550 & 0.655 & 0.521 \\
 & $\pm$0.112 & $\pm$0.125 & $\pm$0.216 & $\pm$0.178 \\
DP ($\varepsilon$=8) & 0.707 & 0.699 & 0.777 & 0.676 \\
 & $\pm$0.075 & $\pm$0.075 & $\pm$0.148 & $\pm$0.124 \\
VIB-only & 0.770 & 0.751 & 0.832 & 0.784 \\
 & $\pm$0.074 & $\pm$0.107 & $\pm$0.181 & $\pm$0.197 \\
StandardMINE & 0.714 & 0.716 & 0.815 & 0.664 \\
 & $\pm$0.082 & $\pm$0.093 & $\pm$0.122 & $\pm$0.146 \\
TimeAwareMINE & 0.782 & 0.740 & 0.756 & 0.863 \\
 & $\pm$0.081 & $\pm$0.097 & $\pm$0.162 & $\pm$0.127 \\
\textbf{InfoShield} & \textbf{0.784} & \textbf{0.774} & \textbf{0.776} & \textbf{0.853} \\
 & $\pm$\textbf{0.097} & $\pm$\textbf{0.078} & $\pm$\textbf{0.125} & $\pm$\textbf{0.139} \\
\hline
\end{tabular}
}
\end{table}

\begin{table}[htb]
\centering
\caption{Privacy Protection Against Attribute Inference Attacks (5-fold CV)}
\label{tab:privacy}
{\footnotesize
\begin{tabular}{lcc}
\hline
Method & Gender Acc (\%) & Age Acc (\%) \\
\hline
Raw Features (No Privacy) & 92.6 $\pm$ 4.1 & 55.7 $\pm$ 10.8 \\
\hline
Normal & 77.8 $\pm$ 4.2 & 43.9 $\pm$ 8.8 \\
VIB-only & 61.3 $\pm$ 17.3 & 45.6 $\pm$ 25.4 \\
DP ($\varepsilon$=1) & 59.4 $\pm$ 16.7 & 42.0 $\pm$ 10.4 \\
DP ($\varepsilon$=8) & 74.4 $\pm$ 14.1 & 41.7 $\pm$ 9.2 \\
StandardMINE & 54.3 $\pm$ 13.6 & 43.5 $\pm$ 8.8 \\
TimeAwareMINE & 62.2 $\pm$ 19.5 & 39.7 $\pm$ 10.7 \\
\textbf{InfoShield} & \textbf{55.5 $\pm$ 16.9} & \textbf{30.3 $\pm$ 14.6} \\
\hline
\end{tabular}
}
\end{table}

\noindent\textbf{Statistical Significance for Clinical Deployment.} Paired t-tests confirm that InfoShield significantly outperforms all baselines except the Normal oracle in diagnostic accuracy for depression classification ($p < 0.05$), validating clinical feasibility. For privacy protection, InfoShield achieves significantly lower gender and age inference compared to Normal, VIB-only, and DP baselines ($p < 0.05$), demonstrating statistically robust privacy guarantees suitable for healthcare applications.

\textbf{Framework Synergy Analysis.} InfoShield demonstrates synergistic effects. TimeAwareMINE alone achieves 62.2\% gender and 39.7\% age inference, while InfoShield improves to 55.5\% and 30.3\%---a 6.7pp and 9.4pp improvement, respectively. StandardMINE provides better gender privacy (54.3\%) than TimeAwareMINE (62.2\%), but TimeAwareMINE offers superior age privacy (39.7\% vs. 43.5\%) and better utility (F1: 0.782 vs. 0.714), validating temporal-aware design benefits for sequential speech.

\textbf{Component Ablation Analysis.} Table~\ref{tab:utility} and~\ref{tab:privacy} reveal how each component contributes: (1) \textbf{VIB-only}: Compression provides modest privacy gains (Gender: 61.3\% vs. 77.8\% Normal) through implicit information filtering. (2) \textbf{StandardMINE}: Direct privacy optimization achieves strong gender privacy (54.3\%) but suffers utility degradation (F1: 0.714) and weaker age protection (43.5\%). (3) \textbf{TimeAwareMINE}: Temporal-aware design improves utility (F1: 0.782) and age privacy (39.7\%) vs. StandardMINE. (4) \textbf{InfoShield}: Complete framework achieves optimal balance (Gender: 55.5\%, Age: 30.3\%, F1: 0.784), demonstrating synergistic benefits.

\textbf{Tradeoff and Failure Analysis.} Varying the privacy weight $\gamma$ reveals a monotonic relationship: increasing $\gamma$ reduces gender inference (62.2\% $\to$ 55.5\% $\to$ 48.3\%) and age inference (39.7\% $\to$ 30.3\% $\to$ 27.1\%), while depression F1 degrades (0.782 $\to$ 0.784 $\to$ 0.721). The selected $\gamma=0.01$ achieves age inference near random chance (33.3\%) with 6\% utility loss. However, we observe high variance across folds ($\pm$15-20\% for some methods), likely due to the small dataset size. Age protection proves more challenging than gender—the three-class age task has higher random baseline, and adjacent age groups (Young vs. Middle) are frequently confused by the attacker.

\subsection{Information-Theoretic vs. Differential Privacy}

InfoShield demonstrates clear superiority over DP methods. Compared to strong DP ($\varepsilon$=1), we achieve better privacy (Gender: 55.5\% vs. 59.4\%, Age: 30.3\% vs. 42.0\%) with higher utility (F1: 0.784 vs. 0.568). Even against weaker DP ($\varepsilon$=8), we provide superior privacy with comparable utility (F1: 0.784 vs. 0.707).

The key difference: DP's global noise injection affects all learned features indiscriminately, including diagnostically relevant patterns. Our targeted MI minimization selectively removes sensitive information through principled optimization, guided by TimeAwareMINE's accurate quantification. This explains our 38\% utility improvement over strong DP while maintaining superior privacy.

\subsection{Multi-Attribute Privacy Analysis}

Our multi-attribute evaluation validates comprehensive protection across diverse demographics. Against Transformer-based attacks, InfoShield achieves substantial privacy gains: gender inference drops to 55.5\% (22.3pp improvement from Normal) and age inference to 30.3\% (13.6pp improvement, below 33\% random chance).

These improvements demonstrate attribute-specific effectiveness: gender protection reaches near-random levels, while age protection significantly exceeds random chance, suggesting TimeAwareMINE's temporal-aware design provides superior protection for temporal-dependent attributes. Consistent performance across binary (gender) and multi-class (age) tasks validates generalizability. InfoShield outperforms VIB-only (61.3\% vs. 55.5\% gender, 45.6\% vs. 30.3\% age) while achieving superior diagnostic performance, confirming explicit MI minimization benefits over implicit compression.

\section{Conclusion}

This paper presents InfoShield, a unified information-theoretic framework addressing privacy concerns in speech-based mental health technologies. By integrating VIB compression with TimeAwareMINE, we achieve utility-privacy balance for healthcare applications.

\textbf{Summary of Contributions.} Our work delivers three key advancements: (1) \textbf{TimeAwareMINE}: We address the temporal-static misalignment in sequential speech by introducing cross-modal attention. This mechanism yields superior outcomes over StandardMINE, improving utility (F1: 0.782 vs. 0.714) while tightening age privacy (39.7\% vs. 43.5\%). (2) \textbf{Unified InfoShield Framework}: By synergizing VIB compression with our targeted MI minimization, we achieve a robust privacy-utility trade-off (Gender: 55.5\%, Age: 30.3\%, F1: 0.784) that surpasses individual components. (3) \textbf{Empirical Validation}: On the Androids Corpus, InfoShield successfully suppresses attribute leakage---reducing gender inference from 92.6\% to 55.5\% and age inference from 55.7\% to 30.3\%---with only a marginal 6\% utility cost, thereby outperforming differential privacy baselines and the prior SOTA~\cite{alsarrani2022thin} (F1: 0.784 vs. 0.723).

\textbf{Limitations and Future Work.} The statistical power of our findings is constrained by the small sample size (118 speakers) and single language (Italian). While this study demonstrates feasibility, future work must validate robustness on larger, multilingual cohorts and extend protection to other sensitive attributes under stronger threat models.

\section{Disclosure of Generative AI Use}
\label{sec:ai-disclosure}

Generative AI tools were used solely for: (1) text polishing and language refinement to improve clarity; (2) code efficiency optimization suggestions; and (3) partial debugging assistance. All core research contributions were completed by the authors, including idea conception, initial drafting, related work analysis, code implementation, and data preprocessing. The authors take full responsibility for the content of this paper.

\bibliographystyle{IEEEtran}
\bibliography{mybib}

@article{cheng2020club,
  title={CLUB: A Contrastive Log-ratio Upper Bound of Mutual Information},
  author={Cheng, Pengyu and Hao, Weituo and Dai, Shuyang and Liu, Jiachang and Gan, Zhe and Carin, Lawrence},
  journal={Proceedings of the 37th International Conference on Machine Learning},
  pages={1779--1788},
  year={2020},
  organization={PMLR}
}

@inproceedings{ravanelli2019learning,
  title={Learning Speaker Representations with Mutual Information},
  author={Ravanelli, Mirco and Bengio, Yoshua},
  booktitle={Interspeech 2019},
  pages={1153--1157},
  year={2019}
}

@article{belghazi2018mine,
  title={Mine: mutual information neural estimation},
  author={Belghazi, Mohamed Ishmael and Baratin, Aristide and Rajeswar, Sai and Ozair, Sherjil and Bengio, Yoshua and Courville, Aaron and Hjelm, R Devon},
  journal={arXiv preprint arXiv:1801.04062},
  year={2018}
}

@article{tao2023androids,
  title={The androids corpus: A new publicly available benchmark for speech based depression detection},
  author={Tao, Fuxiang and Esposito, Anna and Vinciarelli, Alessandro},
  journal={Depression},
  volume={47},
  pages={11--9},
  year={2023}
}

@article{de2024probing,
  title={Probing mental health information in speech foundation models},
  author={de Gennes, Marc and Lesage, Adrien and Denais, Martin and Cao, Xuan-Nga and Chang, Simon and Van Remoortere, Pierre and Dakhlia, Cyrille and Riad, Rachid},
  journal={arXiv preprint arXiv:2409.19042},
  year={2024}
}

@inproceedings{poole2019variational,
  title={On variational bounds of mutual information},
  author={Poole, Ben and Ozair, Sherjil and Van Den Oord, Aaron and Alemi, Alex and Tucker, George},
  booktitle={International conference on machine learning},
  pages={5171--5180},
  year={2019},
  organization={PMLR}
}

@inproceedings{abadi2016deep,
   title={Deep Learning with Differential Privacy},
   url={http://dx.doi.org/10.1145/2976749.2978318},
   DOI={10.1145/2976749.2978318},
   booktitle={Proceedings of the 2016 ACM SIGSAC Conference on Computer and Communications Security},
   publisher={ACM},
   author={Abadi, Martin and Chu, Andy and Goodfellow, Ian and McMahan, H. Brendan and Mironov, Ilya and Talwar, Kunal and Zhang, Li},
   year={2016},
   month=oct, collection={CCS'16} }

@inproceedings{snyder2017deep,
  title={Deep neural network embeddings for text-independent speaker verification},
  author={Snyder, David and Garcia-Romero, Daniel and Povey, Daniel and Khudanpur, Sanjeev},
  booktitle={Proc. Interspeech 2017},
  pages={999--1003},
  year={2017},
  doi={10.21437/Interspeech.2017-620}
}

@article{fang2019speaker,
  title={Speaker anonymization using x-vector and neural waveform models},
  author={Fang, Fuming and Wang, Xin and Yamagishi, Junichi and Echizen, Isao and Todisco, Massimiliano and Evans, Nicholas and Bonastre, Jean-Francois},
  journal={arXiv preprint arXiv:1905.13561},
  year={2019}
}

@inproceedings{fan2019adv,
  title={Adversarial-Based Training for ASR-Robust Speaker Verification and Voice ID-Hiding},
  author={Fan, Long and Sgroi, Aili and St-Hilaire, Marc and Cory, Michael and Du, Jun and Glass, James},
  booktitle={Proc. Interspeech 2019},
  pages={2200--2204},
  year={2019},
  doi={10.21437/Interspeech.2019-2165}
}

@inproceedings{alsarrani2022thin,
  title={Thin slices of depression: Improving depression detection performance through data segmentation},
  author={Alsarrani, Rawan and Esposito, Anna and Vinciarelli, Alessandro},
  booktitle={ICASSP 2022-2022 IEEE International Conference on Acoustics, Speech and Signal Processing (ICASSP)},
  pages={6257--6261},
  year={2022},
  organization={IEEE}
}

@inproceedings{tishby2000information,
  title={The information bottleneck method},
  author={Tishby, Naftali and Pereira, Fernando C and Bialek, William},
  booktitle={Proceedings of the 37th Annual Allerton Conference on Communication, Control, and Computing},
  pages={368--377},
  year={2000},
  organization={Citeseer}
}

@inproceedings{dwork2006differential,
  title={Differential privacy},
  author={Dwork, Cynthia},
  booktitle={International colloquium on automata, languages, and programming},
  pages={1--12},
  year={2006},
  organization={Springer}
}

@article{who2022mental,
  title={World mental health report: Transforming mental health for all},
  author={{World Health Organization}},
  journal={World Health Organization},
  year={2022},
  url={https://www.who.int/publications/i/item/9789240049338}
}

@article{cummins2018speech,
  title={Speech analysis for health: Current state-of-the-art and the increasing impact of deep learning},
  author={Cummins, Nicholas and Baird, Alice and Schuller, Bj{\"o}rn W},
  journal={Methods},
  volume={151},
  pages={41--54},
  year={2018},
  publisher={Elsevier}
}

@article{low2020voice,
  title={Voice-based detection of Alzheimer's disease using machine learning},
  author={Low, Daniel M and Bentley, Kate H and Ghosh, Satrajit S},
  journal={PLoS One},
  volume={15},
  number={12},
  pages={e0244285},
  year={2020},
  publisher={Public Library of Science San Francisco, CA USA}
}

@article{sentencebert,
  author       = {Nils Reimers and
                  Iryna Gurevych},
  title        = {Sentence-BERT: Sentence Embeddings using Siamese BERT-Networks},
  journal      = {CoRR},
  volume       = {abs/1908.10084},
  year         = {2019},
  url          = {http://arxiv.org/abs/1908.10084},
  eprinttype    = {arXiv},
  eprint       = {1908.10084},
  bibsource    = {dblp computer science bibliography, https://dblp.org}
}

@article{pelikan2023federated,
  title={Federated learning with differential privacy for end-to-end speech recognition},
  author={Pelikan, Martin and Azam, Sheikh Shams and Feldman, Vitaly and Silovsky, Jan and Talwar, Kunal and Likhomanenko, Tatiana and others},
  journal={arXiv preprint arXiv:2310.00098},
  year={2023}
}

@article{vecino2025universal,
  title={Universal semantic disentangled privacy-preserving speech representation learning},
  author={Vecino, Biel Tura and Maji, Subhadeep and Varier, Aravind and Bonafonte, Antonio and Valles, Ivan and Owen, Michael and R{\"a}del, Leif and Strimel, Grant and Feyisetan, Seyi and Chicote, Roberto Barra and others},
  journal={arXiv preprint arXiv:2505.13085},
  year={2025}
}

@inproceedings{safeear2024,
  title={Safeear: Content privacy-preserving audio deepfake detection},
  author={Li, Xinfeng and Li, Kai and Zheng, Yifan and Yan, Chen and Ji, Xiaoyu and Xu, Wenyuan},
  booktitle={Proceedings of the 2024 on ACM SIGSAC Conference on Computer and Communications Security},
  pages={3585--3599},
  year={2024}
}

@book{cover1999elements,
  title={Elements of information theory},
  author={Cover, Thomas M},
  year={1999},
  publisher={John Wiley \& Sons}
}

\end{document}